\def\year{2018}\relax
\documentclass[letterpaper]{article} %DO NOT CHANGE THIS
\usepackage{aaai18}  %Required
\usepackage{times}  %Required
\usepackage{helvet}  %Required
\usepackage{courier}  %Required
\usepackage{url}  %Required
\usepackage{graphicx}  %Required
\usepackage[utf8]{inputenc} % allow utf-8 input
\usepackage[T1]{fontenc}    % use 8-bit T1 fonts

\usepackage{url}            % simple URL typesetting
\usepackage{booktabs}       % professional-quality tables
\usepackage{amsfonts}       % blackboard math symbols
\usepackage{nicefrac}       % compact symbols for 1/2, etc.
\usepackage{microtype}      % microtypography
\usepackage{mathtools, cuted}
\usepackage{macros}
\usepackage{multicol}
\usepackage{enumitem}
\usepackage{subcaption}

\newcommand{\kad}{\ensuremath{\operatorname{KAD}}\xspace}
\newcommand{\omkad}{\ensuremath{\operatorname{OM-KIID}}\xspace}
\newcommand{\omrr}{\ensuremath{\operatorname{OM-RR-KAD}}\xspace}
\newcommand{\sos}{\ensuremath{\operatorname{SOS}}\xspace}
\newcommand{\uralg}{\ensuremath{\operatorname{UR-ALG}}\xspace}
\newcommand{\greedy}{\ensuremath{\operatorname{GREEDY}}\xspace}
\newcommand{\sclp}{\ensuremath{\operatorname{ALG-SC-LP}}\xspace}
\newcommand{\egreedy}{\ensuremath{\operatorname{\epsilon-GREEDY}}\xspace}
\newcommand{\re}{C}  % the random variable associated with each $e$

\newcommand{\nr}[1]{E_{#1}}  % the set of neighboring edges incident to either u or v
\newcommand{\SF}{\operatorname{SF}\xspace}
\newcommand{\NALG}{\ensuremath{\operatorname{NADAP}}\xspace}
\newcommand{\adap}{\ensuremath{\operatorname{ADAP}}\xspace}
\newcommand{\citet}[1]{\citeauthor{#1}~\shortcite{#1}}

\title{Allocation Problems in Ride-Sharing Platforms: Online Matching with Offline Reusable Resources}
\author{
	John P. Dickerson\\
	University of Maryland, College Park, USA\\
	\texttt{john@cs.umd.edu} \\
	 \And
	Karthik A. Sankararaman\thanks{Supported in part by NSF Awards CNS 1010789 and CCF 1422569}\\
	University of Maryland, College Park, USA\\
	\texttt{kabinav@cs.umd.edu} \\
	\AND
	Aravind Srinivasan\thanks{Supported in part by NSF Awards CNS-1010789, CCF-1422569 and CCF-1749864, and by research awards from Adobe, Inc}\\
	University of Maryland, College Park, USA\\
	\texttt{srin@cs.umd.edu} \\
	\And
	Pan Xu\thanks{Supported in part by NSF Awards CNS 1010789 and CCF 1422569}\\
	University of Maryland, College Park, USA\\
	\texttt{panxu@cs.umd.edu} \\
}

\begin{document}

\maketitle

\begin{abstract}
	Bipartite matching markets pair agents on one side of a market with agents, items, or contracts on the opposing side.  Prior work addresses online bipartite matching markets, where agents arrive over time and are dynamically matched to a known set of disposable resources.  In this paper, we propose a new model, \emph{Online Matching with (offline) Reusable Resources under Known Adversarial Distributions (\omrr)}, in which resources on the offline side are \emph{reusable} instead of disposable; that is, once matched, resources become available again at some point in the future. We show that our model is tractable by presenting an \LP-based adaptive algorithm that achieves an online competitive ratio of $\frac{1}{2}-\epsilon$ for any given $\epsilon>0$. We also show that no non-adaptive algorithm can achieve a ratio of $\frac{1}{2}+o(1)$ based on the same benchmark LP. Through a data-driven analysis on a massive openly-available dataset, we show our model is robust enough to capture the application of taxi dispatching services and ride-sharing systems. We also present heuristics that perform well in practice.

\end{abstract}
\setcounter{page}{1}
\pagenumbering{arabic}

%\input{1_introduction}
%%%%% Introduction %%%%%

\section{Introduction}\label{sec:intro}
In bipartite matching problems, agents on one side of a market are paired with agents, contracts, or transactions on the other.  Classical matching problems---assigning students to schools, papers to reviewers, or medical residents to hospitals---take place in a static setting, where all agents exist at the time of matching, are simultaneously matched, and then the market concludes.   In contrast, many matching problems are dynamic, where one side of the market arrives in an \emph{online} fashion and is matched sequentially to the other side.

Online bipartite matching problems are primarily motivated by Internet advertising. In the basic version of the problem, we are given a bipartite graph $G = (U, V, E)$ where $U$ represents the offline vertices (advertisers) and $V$ represents the online vertices (keywords or impressions). There is an edge $e=(u,v)$ if advertiser $u$ bids for a keyword $v$. When a keyword $v$ arrives, a central clearinghouse must make an instant and irrevocable decision to either reject $v$ or assign $v$ to one of its ``neighbors'' (\ie set of incident edges) $u$ and obtain a profit $w_e$ for the match $e=(u,v)$. When an advertiser $u$ is matched, it is no longer available for matches with other keywords (in the most basic case) or its budget is reduced. The goal is to design an efficient online algorithm such that the expected total weight (profit) of the matching obtained is maximized. Following the seminal work of~\citet{kvv}, there has been a large body of research on related variants (overviewed by~\citet{mehta2012online}). One particular flavor of problems is online matching with known identical independent distributions (\omkad)~\cite{feldman2009online,haeupler2011online,manshadi2012online,jaillet2013online,brubach2016new}. In this flavor, agents arrive over $T$ rounds, and their arrival distributions are assumed to be {\em identical and independent} over all $T$ rounds; additionally, this distribution is known to the algorithm beforehand.

%more information regarding the online arrival process: we have $T$ rounds and during each round the arrival distributions are assumed to be independent and known in advance. 

Apart from the Internet advertising application, online bipartite matching models have been used to capture a wide range of online resource allocation and scheduling problems. Typically we have an offline and an online party representing, respectively, the service providers (SP)  and online users; once an online user arrives, we need to match it to an offline SP immediately. In many cases, the service is \emph{reusable} in the sense that once an SP is matched to a user, it will be gone for some time, but will then rejoin the system afterwards. Besides that, in many real settings the arrival distributions of online users do change from time to time (\ie they are not \iid). Consider the following motivational examples.

{\bf Taxi Dispatching Services and Ride-Sharing Systems.}  Traditional taxi services and rideshare systems like Uber and Didi Chuxing match drivers to would-be riders~\cite{tong2016online,Lowalekar16:Online,lee2004taxi,seow2010collaborative}.  Here, the offline SPs are different vehicle drivers. 
Once an online request (potential rider) arrives, the system matches it to a nearby driver instantly such that the rider's waiting time is minimized. In most cases, the driver will rejoin the system and can be matched again once she finishes the service. Additionally, the arrival rates of requests changes dramatically across the day. Consider the online arrivals during the peak hours and off-peak hours for example: the arrival rates in the former case can be much larger than the latter.

{\bf Organ Allocation.} Chronic kidney disease affects tens of millions of people worldwide at great societal and monetary cost~\cite{Neuen13:Global,Saran15:US}.  Organ donation---either via a deceased or living donor---is a lifesaving alternative to organ failure.  In the case of kidneys, a donor organ can last up to $15$ years in a patient before failing again.  Various nationwide organ donation systems exist and operate under different ethical and logistical constraints~\cite{Bertsimas13:Fairness,Dickerson15:FutureMatch,Mattei17:Mechanisms}, but all share a common online structure: the offline party is the set of patients (who reappear every $5$ to $15$ years based on donor organ longevity), and the online party is the set of donors or donor organs, who arrive over time.

Similar scenarios can be seen in other areas such as wireless network connection management (SPs are different wireless access points)~\cite{yiu2008capacity} and online cloud computing service scheduling~\cite{miller2008cloud,younge2010efficient}. Inspired by the above applications, we generalize the model of  \omkad in the following two ways.  

 \xhdr{Reusable Resources.} Once we assign $v$ to $u$, $u$ will rejoin the system after $\re_e$ rounds with $e=(u,v)$, where $C_e  \in \{0,1,\ldots, T\} $ is an integral random variable with known distribution. In this paper, we call $\re_e$ the \emph{occupation time} of $u$ w.r.t. $e$. In fact, we show that our setting can directly be extended to the case when $\re_e$ is time sensitive: when matching $v$ to $u$ at time $t$, $u$ will rejoin the system after $\re_{e,t}$ rounds. This extension makes our model adaptive to nuances in real-world settings. For example, consider the taxi dispatching or ride-sharing service: the occupation time of a driver $u$ from a matching with an online user $v$ does depend on both the user type of $v$ (such as destination) and the time when the matching occurs (peak hours can differ significantly from off-peak hours).
 
 \xhdr{Known Adversarial Distributions (\kad).} Suppose we have $T$ rounds and that for each round $t \in [T]$ \footnote{Throughout this paper, we use $[N]$ to denote the set $\{1,2,\ldots,N\}$, for any positive integer $N$.}, a vertex $v$ is sampled from $V$ according to an arbitrary known distribution $\mathcal{D}$ where the marginal for $v$ is $\{p_{v,t}\}$ such that $\sum_{v \in V} p_{v,t} \le 1$ for all $t$. Also, the arrivals at different times are independent (and according to these given distributions). The setting of KAD was introduced by~\cite{alaei2012online,alaei2013online} and is known as Prophet Inequality matching.
 
 We call our new model Online Matching with (offline) Reusable Resources under Known Adversarial Distributions (\omrr, henceforth). 
Note that the \omkad model can be viewed as a special case when $\re_e$ is a constant (with respect to $T$) and $\{p_{v,t}|v\in V\}$ are the same for all $t \in [T]$. 

\xhdr{Competitive Ratio.} Let $\mathbb{E}[\ALG(\mathcal{I}, \mathcal{D})]$ denote the expected value obtained by an algorithm $\ALG$ on an input $\mathcal{I}$ and arrival distribution $\mathcal{D}$. Let $\mathbb{E}[\OPT(\mathcal{I})]$ denote the expected \emph{offline optimal}, which refers to the optimal solution when we are allowed to make choices after observing the entire sequence of online arrival vertices. Then, competitive ratio is defined as $\min_{\mathcal{I}, \mathcal{D}} \frac{\mathbb{E}[\ALG(\mathcal{I}, \mathcal{D})}{\mathbb{E}[\OPT(\mathcal{I})]}$. It is a common technique to use an \LP optimal value to upper bound the $\mathbb{E}[\OPT(\mathcal{I})]$ (called the benchmark \LP) and hence get a valid lower bound on the resulting competitive ratio.

\subsection{Our Contributions}\label{sec:intro-contributions}
First, we propose a new model of \omrr to capture a wide range of real-world applications related to online scheduling, organ allocation, rideshare dispatch, among others. We claim that this model is tractable enough to obtain good algorithms with theoretically provable guarantees and general enough to capture many real-life instances. Our model assumptions take a significant step forward from the usual assumptions in the online matching literature where the offline side is assumed to be \emph{single-use} or \emph{disposable}.
This leads to a larger range of potential applications which can be modeled by online matching.

Second, we show how this model can be \emph{cleanly} analyzed under a theoretical framework. We first construct a linear program (\LP henceforth)~\LP~\eqref{LP:offline-a} which we show is a valid upper-bound on the expected offline optimal (note the latter is hard to characterize). Next, we propose an efficient algorithm that achieves a competitive ratio of $\frac{1}{2}-\epsilon$ for any given $\epsilon>0$. This algorithm solves the $\LP$ and obtains an optimal fractional solution. It uses this optimal solution as a guide in the online phase. Using Monte-Carlo simulations (called simulations henceforth), and combining with this optimal solution, our algorithm makes the online decisions. In particular, Theorem~\ref{thm:1} describes our first theoretical results formally.

\begin{theorem}\label{thm:1}
\LP~\eqref{LP:offline-a} is a valid benchmark for \omrr. There exists an online algorithm, based on \LP~\eqref{LP:offline-a}, achieving an online competitive ratio of $\frac{1}{2}-\epsilon$ for any given $\epsilon>0$.
\end{theorem}

Third, we show that our simple algorithm is nearly optimal among all non-adaptive algorithms. We show that no non-adaptive algorithm can achieve a competitive ratio better than $\frac{1}{2}$ if using \LP~\eqref{LP:offline-a} as the benchmark. Formally, Theorem~\ref{thm:2} states this result.

\begin{theorem}\label{thm:2}
No non-adaptive algorithm, based on benchmark \LP~\eqref{LP:offline-a}, can achieve a competitive ratio better than $\frac{1}{2}+o(1)$ \footnote{$o(1)$ is a vanishing term when both of $C_e$ and $T/C_e$ are sufficiently large.} even when all $C_e$ are constants.
\end{theorem}

Finally, through a data-driven analysis on a massive openly available dataset we show that our model is robust enough to capture the setting of taxi hailing/sharing at least. Additionally, we provide certain \emph{simpler} heuristics which also give good performance. Hence, we can \emph{combine} these theoretically grounded algorithms with such heuristics to obtain further improved ratios in practice.  Section~\ref{sec:experiments} provides a detailed qualitative and quantitative discussion.

%\input{2_related}
%%%%%%%%%%%%%% Related %%%%%%%%%%%%
\subsection{Other Related Work}\label{sec:intro-related}
In addition to the arrival assumptions of KIID and KAD, there are several other important, well-studied variants of online matching problems. Under \emph{adversarial ordering}, an adversary can arrange the arrival order of all items in an arbitrary way (e.g., online matching~\cite{kvv,sun2016near} and AdWords~\cite{buchbinder2007online,mehta2007adwords}). 
Under a \emph{random arrival order}, all items arrive in a random permutation order (e.g., online matching~\cite{mahdian2011online} and AdWords~\cite{goel2008online}). Finally, under \emph{unknown distributions}, in each round, an item is sampled from a fixed but unknown distribution. (e.g.,~\cite{devanur2011near}). For each of the above categories, we list only a few examples considered under that setting. For a more complete list, please refer to the book by~\citet{mehta2012online}.

%Even though our model is directly inspired  from lots of applicaitons related to online matching,

Despite the fact that our model is inspired by online bipartite matching, it also overlaps with stochastic online scheduling problems (\sos)~\cite{megow2004stochastic,megow2006models,skutella2016unrelated}. We first restate our model 
in the language of \sos: we have $|U|$ {\it nonidentical } parallel machines and $|V|$ jobs; at every time-step a single job $v$ is sampled from $V$ with probability $p_{v,t}$; the jobs have to be assigned immediately after its arrival; additionally each job $v$ can be processed {\it non-preemptively} on a specific subset of machines; once we assign $v$ to $u$, we get a profit of $w_e$ and $u$ will be occupied for $C_e$ rounds with $e=(u,v)$, where $C_e$ is a random variable with known distribution. Observe that the key difference between our model and \sos is in the objective: the former is to maximize the expected profit from the completed jobs, while the latter is to minimize the total or the maximum completion time of all jobs.

%\input{5_main_model}

%%%%%%%%% Main Model %%%%%%%%%%
\section{Main Model}
\label{sec:main-model}
In this section, we present a formal statement of our main model. Suppose we have a bipartite graph $G=(U,V, E)$ where $U$ and $V$ represent offline and online parties respectively. 
 We have a finite time horizon $T$ (known beforehand) and for each time $t \in [T]$, a vertex $v$ will be sampled (we use the term $v$ \emph{arrives}) from a known probability distribution $\{p_{v,t}\}$ such that $\sum_{v \in V} p_{v,t} \le 1$\footnote{Thus, with probability $1-\sum_{v \in V} p_{v,t}$, none of the vertices from $V$ will arrive at $t$.} (noting that such a choice is made independently for each round $t$). The expected number of times $v$ arrives across the $T$ rounds, $\sum_{t \in [T]} p_{v, t}$ is called the \emph{arrival rate} for vertex $v$. Once a vertex $v$ arrives, we need to make an {\it irrevocable decision immediately}: either to reject $v$ or assign $v$ to one of its neighbors in $U$. For each $u$, once it is assigned to some $v$, it becomes unavailable for $\re_e$ rounds with $e=(u,v)$, and subsequently rejoins the system. 
Here $\re_e$ is an integral random variable taking values from $\{0,1,\ldots,T\}$ and the distribution is known in advance. Each assignment $e$ is associated with a weight $w_e$ and our goal is to design an online assignment policy such that the total expected weights of all assignments made is maximized. Following prior work, we assume $|V| \gg |U|$ and $T\gg 1$. Throughout this paper, we use edge $e=(u,v)$ and assignment of $v$ to $u$ interchangeably. 

For an assignment $e$, let $x_{e,t}$ be the probability that $e$ is chosen at $t$ in any offline optimal algorithm. For each $u$ (likewise for $v$), let $\nr{u}$ ($\nr{v}$) be the set of neighboring edges incident to $u$ ($v$). We use the LP \eqref{LP:offline-a} as a benchmark to upper bound the offline optimal. We now interpret the constraints. For each round $t$, once an online vertex $v$ arrives, we can assign it to at most one of its neighbors. Thus, we have: if $v$ arrives at $t$, the total number of assignments for $v$ at $t$ is at most 1; if $v$ does not arrive, the total is 0. The LHS  of \eqref{cons:v} is exactly the expected number of assignments made at $t$ for $v$. It should be no more than the prob. that $v$ arrives at $t$, which is the RHS of \eqref{cons:v}. Constraint \eqref{cons:u} is the \emph{most} novel part of our problem formulation. Consider a given $u$ and $t$. In the LHS, the first term (summation over $t' < t$ and $e \in \nr{u}$) refers to the prob. that $u$ is not available at $t$ while the second term (summation over $e \in \nr{u}$) is the prob. that $u$ is assigned to some worker at $t$, which is no larger than prob. $u$ is available at $t$. Thus, the sum of the first term and second term on LHS is no larger than 1.\footnote{We would like to point out that our LP constraint~\eqref{cons:u} on $u$ is inspired by~\citet{ma2014improvements}. The proof is similar to that by~\citet{alaei2012online} and~\citet{alaei2013online}. 
} This argument implies that the \LP forms a valid upper-bound on the offline optimal solution and hence we have Lemma~\ref{lem:lp}.

\begin{figure*}{!h}
\begin{alignat}{2}
\label{LP:offline-a}
\text{maximize}    &  \sum_{t\in [T]} \sum_{e \in E} w_{e}x_{e,t}  & \\
\text{subject to}  &  \sum_{e \in \nr{v} } x_{e,t} \le  p_{v,t}     &  \qquad \forall v \in V, t\in [T] \label{cons:v}\\
                   & \sum_{t'<t} \sum_{e \in \nr{u} }x_{e,t'} \Pr[C_e>t-t'] +\sum_{e \in \nr{u}} x_{e,t}        \le 1    & \qquad \forall u \in U, t \in [T]\label{cons:u}\\
                   & 0 \le x_{e,t} \le 1   & \forall e \in E, t\in [T] \label{LP:offline-d}
\end{alignat}
\end{figure*}

\begin{lemma}\label{lem:lp}
The optimal value to LP \eqref{LP:offline-a} is a valid upper bound for the offline optimal. 
\end{lemma}

%\input{5_main_alg}
%%%%%%%%% Main Alg %%%%%%%%%%
\section{Simulation-based Algorithm}
\label{sec_main_alg}
		%We present a simulation-based adaptive algorithm. 
		 In this section, we present a simulation-based algorithm. Proofs for Lemma~\ref{lem:sim}, \ref{eqn:hard-2} and \ref{lem:hard-3} can be found in the supplementary material. The main idea is as follows. Let $\vec{x}^*$ denote an optimal solution to LP~\eqref{LP:offline-a}. Suppose we aim to develop an online algorithm achieving a ratio of $\gamma \in [0,1]$. Consider an assignment $e=(u,v)$ when some $v$ arrived at time $t$. Let $\SF_{e,t}$ be the event that $e$ is safe at $t$, \ie $u$ is available at $t$. By simulating the current strategy up to $t$, we can get an estimation of $\Pr[\SF_{e,t}]$, say $\beta_{e,t}$, within an arbitrary small error. Therefore in the case $e$ is safe at $t$, we can sample it with probability $\frac{x^*_{e,t}}{p_{v,t}}\frac{\gamma}{\beta_{e,t}}$, which leads to the fact that $e$ is sampled with probability $\gamma x^*_{e,t}$ unconditionally. Hence, we call any algorithm that satisfies $\gamma \leq \beta_{e, t}$ as \emph{valid}.
 
 The simulation-based attenuation technique has been used previously for other problems, such as stochastic knapsack~\cite{ma2014improvements} and stochastic matching~\cite{adamczyk2015improved}. Throughout the analysis, we assume that we know the exact value of $\beta_{e,t} := \Pr[\SF_{e,t}]$ for all $t$ and $e$. (It is easy to see that the sampling error can be folded into a multiplicative factor of $(1 - \epsilon)$ in the competitive ratio by standard Chernoff bounds and hence, ignoring it leads to a cleaner presentation). The formal statement of our algorithm, denoted by $\adap(\gamma)$, is as follows. For each $v$ and $t$, let $\nr{v,t}$ be the set of {\it safe} assignments for $v$ at  $t$.
 
 %available neighbors of $v$ at $t$ and $\cN_v$ be the set of neighbors of $v$ initially. 

  \begin{algorithm}[h]
\label{alg:adap}
\caption{Simulation-based adaptive algorithm ($\adap (\gamma))$} 
\DontPrintSemicolon
For each time $t$, let $v$ denote the request arriving at time $t$.\;
If $\nr{v,t}=\emptyset$, then reject $v$; otherwise choose $e \in \nr{v,t}$ with prob. $\frac{x^*_{e,t}}{p_{v,t}}\frac{\gamma}{\beta_{e,t}}$ where $e=(u,v)$.
\end{algorithm}

\begin{lemma}\label{lem:sim}
$\adap(\gamma)$ is valid with $\gamma=\frac{1}{2}$.
\end{lemma}

The main Theorem~\ref{thm:1} follows directly from Lemmas~\ref{lem:lp} and~\ref{lem:sim}.

\xhdr{Extension from $C_e$ to $C_{e,t}$.} Consider the case when the occupation time of $u$ from $e$ is sensitive to $t$. In other words, each $u$ will be unavailable for $C_{e,t}$ rounds from the assignment $e=(u,v)$ at $t$. We can accommodate the extension by simply updating the constraints~\eqref{cons:u} on $u$ in the benchmark LP~\eqref{LP:offline-a} to the following. We have that $\forall u \in U, t \in [T]$,

\begin{equation}\label{cons:new_u}
\sum_{t'<t} \sum_{e \in \nr{u} }x_{e,t'} \Pr[C_{e,t'}>t-t'] +\sum_{e \in \nr{u}} x_{e,t}  \le 1   
\end{equation}
The rest of our algorithm remains the same as before. We can verify that (1) LP~\eqref{LP:offline-a} with constraints~\eqref{cons:u} replaced by~\eqref{cons:new_u} is a valid benchmark; (2) $\adap$ achieves a competitive ratio of $\frac{1}{2}-\epsilon$ for any given $\epsilon>0$ for the new model based on the new valid benchmark LP. The modifications to the analysis transfer through in a straightforward way and for brevity we omit the details here.

%\input{5_hardness}
%%%%%%%%%%% Hardness %%%%%%%%%%%

\section{Hardness Result}
\label{sec:hard}
Consider a complete bipartite graph of $G=(U,V,E)$ where $|U|=K$, $|V|=n^2$. Suppose we have $T=n$ rounds and $p_{v,t}=\frac{1}{n^2}$ for each $v$ and $t$. In other words, in each round $t$, each $v$ is sampled uniformly from $V$. For each $e$, let $C_e$ be a constant of $K$, which implies that each $u$ will be unavailable for a constant $K$ rounds after each assignment. Assume all assignments have a uniform weight (\ie $w_e=1$ for all $e$). Split the whole online process of $n$ rounds into $n-K+1$ consecutive windows $\cW=\{W_\ell\}$ such that $W_\ell=\{\ell, \ell+1, \ldots, \ell+K-1\}$ for each $1 \le \ell \le n-K+1$. The benchmark LP~\eqref{LP:offline-a} then reduces to the following.

\begin{alignat}{2}
\label{LP:offline-b}
\text{max}    &  \sum_{t\in [T]} \sum_{e \in E} x_{e,t}   \\
\text{s.t.}  &  \sum_{e \in \nr{v} } x_{e,t} \le  \frac{1}{n^2}     & \ & \quad\forall v \in V, t\in [T] \label{cons:vb}\\
                   & \sum_{t \in W_\ell} \sum_{e \in \nr{u} }x_{e,t}       \le 1    &\ & \quad\forall u \in U, 1\le \ell  \le n-K+1 \label{cons:ub}\\
                   & 0 \le x_{e,t} \le 1                   &\ & \quad\forall e \in E, t\in [T] \label{LP:offline-end-b}
\end{alignat}
 
We can verify that an optimal solution to the above LP is as follows: $x^*_{e,t}= 1/(n^2K)$ for all $e$ and $t$ with the optimal objective value of $n$. We investigate the performance of any optimal non-adaptive algorithm. Notice that the expected arrivals of any $v$ in the full sequence of online arrivals is $1/n$. Thus for any non-adaptive algorithm $\NALG$,  it needs to specify the allocation distribution $\cD_v$ for each $v$ during the first arrival. Consider a given $\NALG$ parameterized by $\{\alpha_{u,v} \in [0,1]\}$ for each $v$ and $u \in \nr{v}$ such that $\sum_{u \in \nr{v}} \alpha_{u,v} \le 1$ for each $v$. In other words, $\NALG$ will assign $v$ to $u$ with probability $\alpha_{u,v}$ when $v$ comes for the first time and $u$ is available. 

Let $\beta_u=\sum_{v \in \nr{u}} \alpha_{u,v}*\frac{1}{n^2}$, which is the probability that $u$ is matched in each round if it is safe at the beginning of that round, when running $\NALG$. Hence, 
\begin{equation*}\label{eqn:hard-1}
\sum_{u \in U} \beta_u = \sum_{u \in U}\sum_{v \in \nr{u}} \alpha_{u,v}*\frac{1}{n^2} =
\sum_{v \in V}\sum_{u \in \nr{v}} \alpha_{u,v}*\frac{1}{n^2} \le 1
\end{equation*}

Consider a given $u$ with $\beta_u$ and let $\gamma_{u,t}$ be the probability that $u$ is available at $t$. Then the expected number of matches of $u$ after the $n$ rounds is $\sum_t \beta_u \gamma_{u,t}$. We have the recursive inequalities on $\gamma_{u,t}$ as in Lemma~\ref{eqn:hard-2}, with $\gamma_{u,t}=1, t=1$.

\begin{lemma}
\label{eqn:hard-2}
$\forall 1<t \le n$, we have	
\[ \gamma_{u,t}+\beta_{u} \sum_{ t-K+1 \le t' <t } \gamma_{u,t'} = 1 \]
\end{lemma}

Note that the \OPT of our benchmark LP is $n$ while the performance of $\NALG$ is $\sum_u \sum_t \beta_u \gamma_{u,t}$. The resulting competitive ratio achieved by an optimal \NALG is captured by the following maximization program.

\begin{equation}\label{eqn:hard-3}
\resizebox{0.91\hsize}{!}{
$
\begin{aligned}
&{\text{max}}
&&\frac{\sum_u  \sum_t \beta_u \gamma_{u,t}}{n} \\
&\text{s.t.} 
&&  \sum_{u \in U} \beta_{u}  \le 1 & \\
&&&\gamma_{u,t}+\beta_{u} \sum_{ t-K+1 \le t' <t } \gamma_{u,t'} = 1
& \forall  1<t\le n, u \in U\\
&&& \beta_u \ge 0, \gamma_{u, 1}=1  & \forall u \in U
\end{aligned}
$
}
\end{equation}

We prove the following Lemma which implies Theorem~\ref{thm:2}.

\begin{lemma}\label{lem:hard-3}
The optimal value to the program~\eqref{eqn:hard-3} is at most $\frac{1}{2-1/K}+o(1)$ when $K=o(n)$. 
\end{lemma}

\xhdr{Unconditional Hardness.} \citet{manshadi2012online} prove that for the online matching problem under known distribution (but disposable offline vertices), no algorithm can achieve a ratio better than $0.823$. Since our setting generalizes this, the hardness results directly apply to our problem as well.

%\input{6_experiments}
%%%%%%%%% Experiments %%%%%%
\section{Experiments}
\label{sec:experiments}
		\begin{figure*}[!ht]
					\minipage{0.32\textwidth}
					\includegraphics[width=\linewidth]{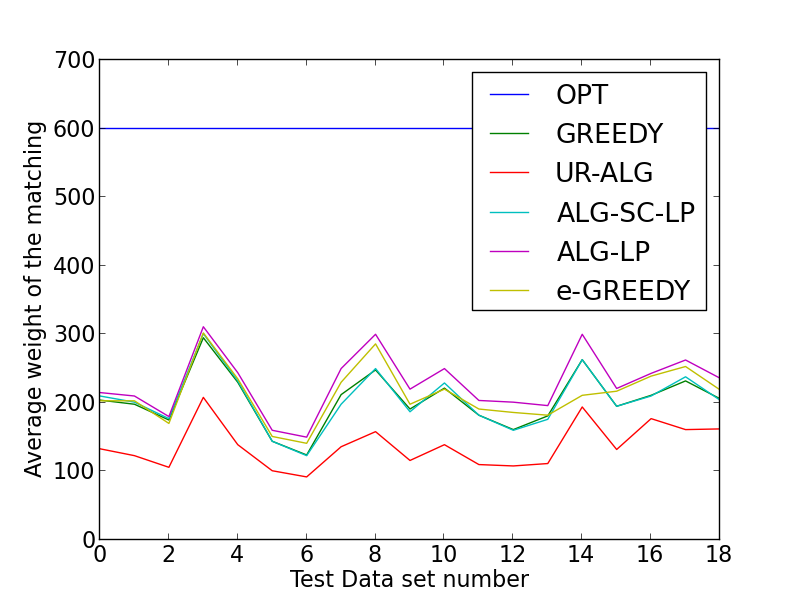}
					\caption{OTD is normal distribution under KIID}
					\label{fig:normaluniform}
					\endminipage\hfill
					\minipage{0.32\textwidth}
					\includegraphics[width=\linewidth]{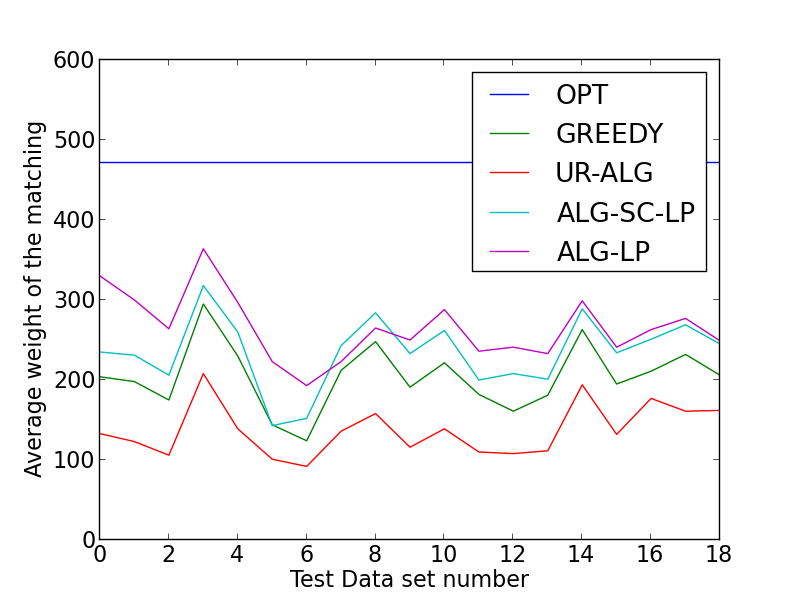}
					\caption{OTD is normal distribution under KAD}\label{fig:normal_pjt}
					\endminipage\hfill
					\minipage{0.32\textwidth}%
					\includegraphics[width=\linewidth]{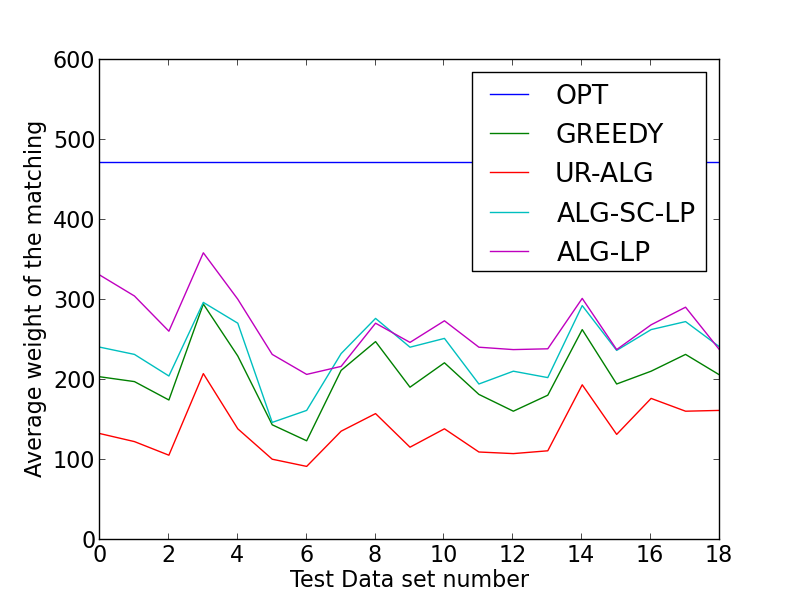}
					\caption{OTD is power law distribution under KAD}\label{fig:powerlaw_pjt}
					\endminipage
				\end{figure*}

				\begin{figure*}[!ht]
					\minipage{0.32\textwidth}
					\includegraphics[width=\linewidth]{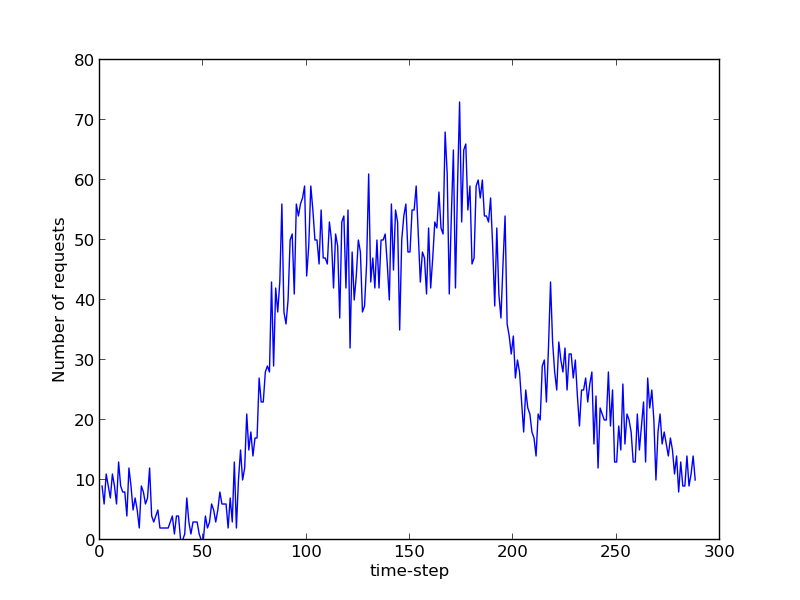}
					\caption{The number of requests of a given type at various time-steps.  x-axis: time-step, y-aixs: number of requests}
					\label{fig:typerequests}
					\endminipage\hfill
					\minipage{0.32\textwidth}
					\includegraphics[width=\linewidth]{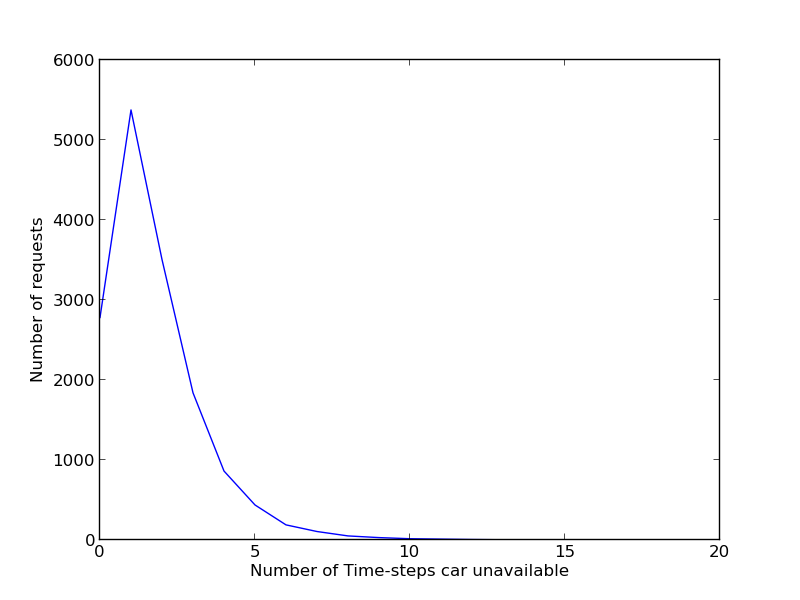}
					\caption{Occupation time distribution of all cars. x-axis: number of time-steps, y-axis: number of requests }\label{fig:reappearDistributionAll}
					\endminipage\hfill
					\minipage{0.32\textwidth}%
					\includegraphics[width=\linewidth]{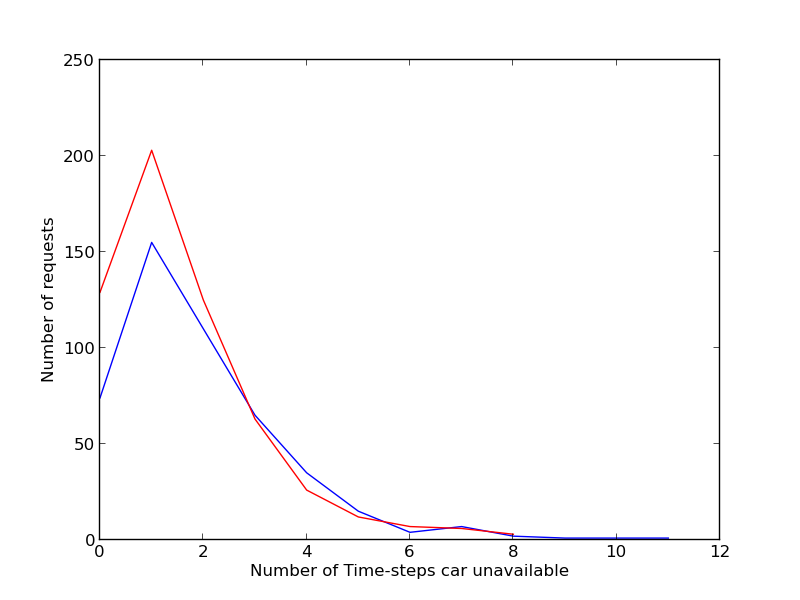}
					\caption{Occupation time distribution of two different cars. x-axis: number of time-steps, y-axis: number of requests}\label{fig:tworeappear}
					\endminipage
				\end{figure*}

	\label{sec:experiments-dataset}
To validate the approaches presented in this paper, we use the New York City yellow cabs dataset,\footnote{\texttt{http://www.andresmh.com/nyctaxitrips/}} which contains the trip records for trips in Manhattan, Brooklyn, and Queens for the year $2013$. The dataset is split into $12$ months. For each month we have numerous records each corresponding to a single trip. Each record has the following structure. We have an anonymized license number which is the primary key corresponding to a car. For privacy purposes a long string is used as opposed to the actual license number. We then have the time at which the trip was initiated, the time at which the trip ended, and the total time of the trip in seconds. This is followed by the starting coordinates (\ie latitude and longitude) of the trip and the destination coordinates of the trip.

			\xhdr{Assumptions.} We make two assumptions specific to our experimental setup. Firstly, we assume that every \emph{car} starts and ends at the same location, for \emph{all} trips that it makes. Initially, we assign every car a location (potentially the same) which corresponds to its \emph{docking} position. On receiving a request, the car leaves from this docking position to the point of pick-up, executes the trip and returns to this docking position. Secondly, we assume that occupation time distributions (OTD) associated with all matches are identically (and independently) distributed, \ie $\{\re_e\}$ follow the same distribution. Note that this is a much stronger assumption than what we made in the model, and is completely inspired by the dataset (see Section~\ref{sec:exp_ana}). We test our model on two specific distributions, namely a \emph{normal} distribution and the \emph{power-law} distribution (see Figure~\ref{fig:reappearDistributionAll}). The docking position of each car and parameters associated with each distribution are all learned from the training dataset (described below in the {\bf Training} discussion).

		\subsection{Experimental Setup}
			For our experimental setup, we randomly select $30$ cabs (each cab is denoted by $u$). We discretize the Manhattan map into cells such that each cell is approximately $4$ miles (increments of $0.15$ degrees in latitude and longitude). For each pair of locations, say $(a,b)$, we create a request \emph{type} $v$, which represents all trips with starting and ending locations falling into $a$ and $b$ respectively. In our model, we have
		$|U|=30$ and $|V| \approx 550$ (variations depending on day to day requests with low variance).
			 We focus on the month of January $2013$. We split the records into $31$ parts, each corresponding to a day of January. We choose a random set of $12$ parts for \emph{training} purposes and use the remaining for \emph{testing} purposes. 

The edge weight $w_e$ on $e=(u,v)$ (\ie edge from a car $u$ to type $v$) is set as a function of two distances in our setup. The first is the trip distance (\ie the distance from the starting location to the ending location of $v$, denoted $L_1$) while the second is the docking distance (\ie the distance from the docking position of $u$ to the starting/ending location of $v$, denoted $L_2$). We set $w_e=\max(L_1-\alpha L_2, 0)$, where $\alpha$ is a parameter capturing the subtle balance between the positive contribution from the trip distance and negative contribution from the docking distance to the final profit. We set $\alpha=0.5$ for the experiments. We consider each single day as the time horizon and set the total number of rounds $T=\frac{24*60}{5}=288$ by discretizing the $24$-hour period into a time-step of $5$ minutes. Throughout this section, we use time-step and round interchangeably.

			\xhdr{Training.} We use the training dataset of $12$ days to learn various parameters. As for the arrival rates $\{p_{v,t}\}$, we count the total number of appearances of each request type $v$ at time-step $t$ in the $12$ parts (denote it by $c_{v,t}$) and set $p_{v,t}=c_{v,t}/12$ under KAD. When assuming KIID, we set  $p_v=p_{v,t}  = (c/12)/T$ (\ie the arrival distributions are assumed the same across all the time-steps for each $v$). The estimation of parameters for the two different occupation time distributions are processed as follows. We first compute the average number of seconds between two \emph{requests} in the dataset (note this was $5$ minutes in the experimental setup). We then assume that each \emph{time-step} of our online process corresponds to a time-difference of this average in seconds. We then compute the sample mean and sample variance of the trip lengths (as number of seconds taken by the trip divided by five minutes) in the $12$ parts. Hence we use the normal distribution obtained by this sample mean and standard deviation as the distribution with which a car is unavailable. 
We assign the docking position of each car to the location (in the discretized space) in which the majority of the requests were initiated (\ie starting location of a request) and matched to this car.
			\subsection{Justifying The Two Important Model Assumptions}
				\label{sec:exp_ana}
				\xhdr{Known Adversarial Distributions.} Figure~\ref{fig:typerequests} plots the number of arrivals of a particular type at various times during the day. Notice the significant increase in the number of requests in the middle of the day as opposed to the mornings and nights. This justified our arrival assumption of KAD which assumes different arrival distributions at different time-steps. Hence the \LP (and the corresponding algorithm) can exploit this vast difference in the arrival rates and potentially obtain improved results compared to the assumption of Known Identical Independent Distributions (KIID). This is confirmed by our experimental results shown in Figures~\ref{fig:normaluniform} and~\ref{fig:normal_pjt}.

				\xhdr{Identical Occupation Time Distribution.}  We assume each car will be available again via an independent and identical random process regardless of the matches it received. The validity of our assumptions can be seen in Figures~\ref{fig:reappearDistributionAll} and~\ref{fig:tworeappear}, where the $x$-axis represents the different occupation time and the $y$-axis represents the corresponding number of requests in the dataset responsible for each occupation time. It is clear that for most requests the occupation time is around 2-3 time-steps and dropping drastically beyond that with a long tail. Figure~\ref{fig:tworeappear} displays occupation times for two representative (we chose two out of the many cars we plotted, at random) cars in the dataset; we see that the distributions roughly coincide with each other, suggesting that such distributions can be learned from historical data and used as a guide for future matches.

			\subsection{Results}
			 Inspired by the experimental setup by \cite{tong2016online,tong2016online_icde}, we run five different algorithms on our dataset. The first algorithm is the $\ALG$-$\LP$. In this algorithm, when a request $v$ arrives, we choose a neighbor $u$ with probability $x^*_{e,t}/p_{v,t}$ with $e=(u,v)$ if $u$ is available. Here $x^*_{e,t}$ is an optimal solution to our benchmark \LP and $p_{v,t}$ is the arrival rate of type $v$ at time-step $t$. The second algorithm is called \sclp. Recall that $E_{v,t}$ is the set of ``safe"
or available assignments with respect to $v$ when the type $v$ arrives at $t$. Let $x_{v,t}= \sum_{e \in E_{v,t}} x^*_{e,t}$. In \sclp, we sample a safe assignment for $v$ with probability $x^*_{e,t}/x_{v,t}$. The next two algorithms are heuristics oblivious to the underlying $\LP$. Our third algorithm is called \greedy which is as follows. When a request $v$ comes, match it to the safe neighbor $u$ with the highest edge weight. Our fourth algorithm is called \uralg which chooses one of the safe neighbors uniformly at random. Finally, we use a combination of $\LP$-oblivious algorithm and $\LP$-based algorithm called \egreedy. In this algorithm when a type $v$ comes, with probability $\epsilon$ we use the greedy choice and with probability $1-\epsilon$ we use the optimal $\LP$ choice. In our algorithm, we optimized the value of $\epsilon$ and set it to $\epsilon = 0.1$. We summarize our results in the following plots. Figures~\ref{fig:normaluniform}, \ref{fig:normal_pjt}, and~\ref{fig:powerlaw_pjt} show the performance of the five algorithms and \OPT (optimal value of the benchmark \LP) under the different assumptions of the OTD (normal or power law) and online arrives (KIID or KAD). In all three figures the x-axis represents test data-set number and the y-axis represents average weight of matching.
	
			\xhdr{Discussion.}
				From the figures, it is clear that both the \LP-based solutions, namely $\ALG$-$\LP$ and \sclp, do better than choosing a free neighbor uniformly at random. Additionally, with distributional assumptions the \LP-based solutions outperform greedy algorithm as well. We would like to draw attention to a few interesting details in these results. Firstly, compared to the \LP optimal solution, our \LP-based algorithms have a competitive ratio in the range of $0.5$ to $0.7$. We believe this is because of our experimental setup. In particular, we have that the rates are high ($>0.1$) only in a few time-steps while in all other time-steps the rates are very close to $0$. This means that it resembles the structure of the \emph{theoretical} worst case example we showed in Section~\ref{sec:hard}. In future experiments, running our algorithms during \emph{peak} periods (where the request rates are significantly larger than $0$) may show that competitive ratios in those cases approach $1$. Secondly, it is surprising that our algorithm is fairly robust to the \emph{actual} distributional assumption we made. In particular, from Figures~\ref{fig:normal_pjt} and~\ref{fig:powerlaw_pjt} it is clear that the difference between the assumption of normal distribution versus power-law distribution for the unavailability of cars is \emph{negligible}. This is important since it might not be easy to learn the \emph{exact} distribution in many cases (\eg cases where the sample complexity is high) and this shows that a close approximation will still be as good.
				
%\input{7_future}
%%%%%%%%%%%% Future %%%%%%%%%%%%%%
\section{Conclusion and Future Directions}
	In this work, we provide a model that captures the application of assignment in ride-sharing platforms. One key aspect in our model is to consider the \emph{reusable} aspect of the offline resources. This helps in modeling many other important applications where agents enter and leave the system \emph{multiple} times (\eg organ allocation, crowdsourcing markets \cite{ho2012online}, and so on). Our work opens several important research directions. The first direction is to generalize the online model to the \emph{batch} setting. In other words, in each round we assume multiple arrivals from $V$. This assumption is useful in crowdsourcing markets (for example) where multiple tasks---but not all---become available at some time. The second direction is to consider a Markov model on the driver starting position. In this work, we assumed that each driver returns to her docking position. However, in many ride-sharing systems, drivers start a new trip from the position of the last-drop off. This leads to a Markovian system on the offline types, as opposed to the assumed static types in the present work. Finally, pairing our current work with more applied stochastic optimization and reinforcement learning approaches would be of practical interest to policymakers running taxi and bikeshare services~\cite{Singhvi15:Predicting,OMahony15:Data,Lowalekar16:Online,Verma17:Augmenting,Ghosh17:Dynamic}.

{\small
\bibliographystyle{aaai}
\bibliography{short_refs}
}

\newpage

\onecolumn
\section{Supplementary Materials}

\subsection{Proof of Lemma~\ref{lem:sim}}

We show by induction on $t$ as follows. When $t=1$, $\beta_{e,t}=1$ for all $e = (u, *)$ and we are done since 
$$\sum_{e \in \nr{v,t}}\frac{x^*_{e,t}}{p_{v,t}}\frac{\gamma}{\beta_{e,t}} \le \sum_{e \in \nr{v}}\frac{x^*_{e,t}}{p_{v,t}} \gamma \le \frac{1}{2}$$

Assume for all $t'<t$, $\beta_{e,t'} \ge 1/2$ and $\adap(\gamma)$ is valid for all rounds before $t$. In other words, each $e$ is made with probability equal to $x^*_{e,t'}*\frac{1}{2}$ for all $t'<t$. Now consider a given $e=(u,v)$. Observe that $e$ is unsafe at $t$ iff $u$ is assigned with some $v'$ at $t'<t$ such that the assignment $e'=(u,v')$ makes $u$ unavailable at $t$. Therefore
$$1-\beta_{e,t}=1-\Pr[\SF_{e,t}]=\sum_{t'<t} \sum_{e \in \nr{u}} \frac{x^*_{e,t'}}{2}\Pr[\re_e>t-t']$$
Thus from the constraints~\eqref{cons:u} in our benchmark LP, we see
$$\beta_{e,t}=1-\sum_{t'<t} \sum_{e \in \nr{u}} \frac{x^*_{e,t'}}{2}\Pr[\re_e>t-t'] \ge \frac{1}{2}+\frac{1}{2}\sum_{e \in \nr{u}} x^*_{e,t} \ge \frac{1}{2}$$

Thus we are done since, $\sum_{e \in \nr{v,t}}\frac{x^*_{e,t}}{p_{v,t}}\frac{\gamma}{\beta_{e,t}} \le \sum_{e \in \nr{v}}\frac{x^*_{e,t}}{p_{v,t}} \le 1$.

\subsection{Proof of Lemma~\ref{eqn:hard-2}}

The inequality for $t=1$ is due to the fact that $u$ is safe at $t=1$. For each time $t>1$, Let $\SF_{u,t}$ be the event that $u$ is safe at $t$ and $A_{u,t}$ be the event that $u$ is matched at $t$. Observe that for each window of time slots $K$, $\{\SF_{u,t}, A_{u,t'}, t-K+1\le t'<t\}$ are disjoint events. Therefore,
	\begin{align*}
	1 & =\Pr[\SF_{u,t}]+\sum_{t-K+1 \le t'<t}\Pr[A_{u,t'}] \\
	& =\gamma_{u,t}+\beta_{u} \sum_{ t-K+1 \le t' <t } \gamma_{u,t}
	\end{align*}

\subsection{Proof of Lemma~\ref{lem:hard-3}}

%Let $H_u \doteq  \frac{\sum_t \beta_u \gamma_{u,t}}{n}$. Notice that for each given $\beta_u$, we can get $H_u$ by recursively solving equations on $\{\beta_u, \gamma_{u,t}\}$. We can show that $H_u$ can be upper bounded by a concave function $g(\beta_u)$. Thus the optimal value to the program~\eqref{eqn:hard-3} is upper bounded by $Kg(1/K)$, which yields the result. 

\begin{proof}
Focus on a given $u$. Notice that $\gamma_{u,t}+\beta_{u} \sum_{ t-K+1 \le t' <t } \gamma_{u,t'} = 1$ for all $1 \le t\le n$. Sum all equations over $t \in [n]$, we have
\begin{eqnarray*}
 \Big(1+\beta_u(K-1)\Big) \sum_{t \in [n]} \gamma_{u,t}&=& n+\beta_u(K-1)\gamma_{u,n}+\beta_u(K-2)\gamma_{u,n-1}+\cdots+ \beta_u\gamma_{u, n-K+2}\\
 & \le &  n+K-1
\end{eqnarray*}
Therefore we have 
$$\sum_{t\in [n]} \gamma_{u,t} \le \frac{n}{1+\beta_u(K-1)}+\frac{K-1}{1+\beta_u(K-1)} \le \frac{n}{1+\beta_u(K-1)}+\frac{1}{\beta_u}$$

Define $H_u \doteq  \sum_t \beta_u \gamma_{u,t}$. From the above analysis, we have
$H_u \le  \frac{n \beta_u}{1+\beta_u(K-1)}+1$. Thus the objective value in the program~\eqref{eqn:hard-3} should be at most
$$ \frac{\sum_u  \sum_t \beta_u \gamma_{u,t}}{n} =\sum_{u \in U} \frac{H_u}{n}
\le \sum_{u\in U}  \frac{ \beta_u}{1+\beta_u(K-1)}+\frac{K}{n}$$

We claim that the optimal value to the program~\eqref{eqn:hard-3} can be upper bounded by the below maximization program:
$$\left\{  \max \sum_{u\in[U]} \frac{\beta_u}{1+\beta_u(K-1)} +\frac{K}{n}:~~ \sum_{u\in U} \beta_u=1, \beta_u \ge 0,\forall u \in U \right\}$$

According to our assumption $K=o(n)$, the second term can be ignored. Let $g(x)=x/(1+x (K-1))$. For any $K \ge 2$, it is a concave function, which implies that maximization of $g$ subject to $\sum_u \beta_u=1$ will be achieved when all $\beta_u=1/K$. The resultant value is $\frac{1}{2-1/K}+o(1)$. Thus we are done.
\end{proof}

%%%%%%%%%%%%%%%%%
\end{document}